\documentclass[journal, letterpaper]{IEEEtran}

\usepackage{graphicx}

\usepackage{url}

\usepackage{amsmath}    
\usepackage{bm}

\usepackage{textgreek}	
\usepackage{listings}
\usepackage{csvsimple}
\usepackage{longtable}
\usepackage{array}
\usepackage{booktabs}
\usepackage{colortbl}
\usepackage{xcolor}
\usepackage{graphicx}

\begin{document}
\title{A Polarization Image Dehazing Method Based on the Principle of Physical Diffusion}
\author{Zhenjun Zhang, Lijun Tang, Hongjin Wang, Lilian Zhang, Yunze He, Yaonan Wang
\thanks{This work was supported in major part by the Natural Science Foundation of Hunan Province under Grant 2021JC0004, and in part by the National Natural Science Foundation of China under Grant 62373138. (Co-corresponding authors: Hongjin Wang and Lilian Zhang.)
}
\thanks{Z. Zhang, L. Tang, H. Wang, Y. He and Y. Wang are with the College of Electrical and Informatic Engineering, Hunan University, Changsha 410082, China (e-mail: zhenjun@hnu.edu.cn; tangaha@hnu.edu.cn; hjwang\_2018@hnu.edu.cn; yhe@vip.163.com; yaonan@hnu.edu.cn).}
\thanks{L. Zhang is with the College of Intelligence Science and Technology, National University of Defense Tech nology, Changsha 410073, China (e-mail: lilianzhang@nudt.edu.cn).}}

\maketitle

\begin{abstract}
Computer vision is increasingly used in areas such
as unmanned vehicles, surveillance systems and remote sensing. However, in foggy scenarios, image degradation leads to loss of
target details, which seriously affects the accuracy and
effectiveness of these vision tasks. Polarized light, due to the fact
that its electromagnetic waves vibrate in a specific direction, is able to resist scattering and refraction effects in complex media more effectively compared to unpolarized light. As a result, polarized light has a greater ability to maintain its polarization
characteristics in complex transmission media and under long- distance imaging conditions. This property makes polarized
imaging especially suitable for complex scenes such as outdoor and underwater, especially in foggy environments, where higher quality images can be obtained. Based on this advantage, we propose an innovative semi-physical polarization dehazing method that does not rely on an external light source. The method simulates the diffusion process of fog and designs a diffusion kernel that corresponds to the image blurriness caused
by this diffusion. By employing spatiotemporal Fourier
transforms and deconvolution operations, the method recovers
the state of fog droplets prior to diffusion and the light inversion
distribution of objects. This approach effectively achieves dehazing and detail enhancement of the scene.
\end{abstract}


\section{Introduction}
	
	\IEEEPARstart{C}{omplex} environments such as bad weather or low light can affect most outdoor optical imaging systems, reducing visibility significantly.
Traditional image processing techniques mainly rely on light intensity and color information, but polarized light can provide a richer description of the scene than these conventional methods\cite{wolff1997polarization}, thus extending the potential of image processing and computer vision applications. Currently, polarization vision technology has demonstrated significant advancements in areas such as object recognition in natural scenes, hull damage detection, and marine biology research. Additionally, atmospheric scattering is one of the main causes of polarization phenomena in nature, and the haze resulting from atmospheric scattering can be effectively mitigated through polarization analysis, thereby enhancing the visibility of long-distance targets\cite{atkinson2021polarized}. In haze scenarios, compared with images acquired by conventional cameras, polarized light images can reduce the loss of details and target information, improve contrast and color reproduction, and are crucial for subsequent tasks such as target detection and recognition, image segmentation, and target tracking. Therefore, the study of polarized image enhancement algorithms holds significant application value in the field of image processing.

Polarization dehazing technology estimates atmospheric light intensity by capturing multiple polarized images of the same scene and performs inversion processing on the degraded scene reflection to obtain a dehazed image. This technology can be categorized into four main methods: differential imaging, the Stokes vector method, polarized differential active imaging, and circularly polarized optical imaging. Among these, differential imaging and the Stokes vector method are classified as passive dehazing techniques, relying solely on the polarization characteristics of natural light for image enhancement. These methods depend on a physical degradation model of the atmosphere\cite{schechner2003polarization}, which states that the total radiation reaching the detector is the sum of direct transmitted light and airlight. The direct transmitted light contains the intensity information of the scene's objects, while airlight, or scattered atmospheric light\cite{narasimhan2000chromatic}, serves as a significant source of interference in the imaging system. Differential imaging technology\cite{schechner2001instant} captures images of the same scene at different polarization angles to estimate atmospheric light intensity through differential operations, thereby recovering dehazed images. Although this method effectively eliminates the influence of airlight, it presents challenges in data collection and involves complex computations. The Stokes vector method\cite{zhang2021review},\cite{liang2014method} leverages the partially polarized nature of atmospheric light, treating direct transmitted light as unpolarized. It separates airlight by utilizing information from the degree of polarization and the angle of polarization, thus recovering direct transmitted light. However, this method requires capturing polarized images from multiple angles and necessitates accurately segmenting the sky region in the image when extracting atmospheric light intensity from that area\cite{namer2005advanced}. Polarization differential active imaging and circularly polarized optical imaging techniques are categorized as active dehazing methods, which utilize artificial polarized light sources to illuminate the scene and analyze the polarization state. Polarization differential active imaging\cite{xu2023research} exploits the characteristic that the polarization state of scattered light changes within a scattering medium, while the polarization state of the target reflection remains relatively stable, thereby preserving the target reflection through differential processing. Although this method demonstrates significant effectiveness in strongly scattering environments, the differential process may result in the loss of image details. Circularly polarized optical imaging technology\cite{hu2018polarimetric} is based on the principle that circularly polarized light maintains its original characteristics while passing through scattering media, whereas the polarization characteristics of linearly polarized light gradually diminish. By detecting the reflected light, this method can recover the true image. Although it exhibits strong anti-scattering capabilities and effectively suppresses noise, it incurs high system design costs and has limited real-time performance. 

Image enhancement techniques have made significant research progress across various fields. Histogram equalization\cite{nithyananda2016review} is a common method that improves overall image quality by adjusting the grayscale distribution to enhance contrast. However, its effectiveness in improving local contrast is limited, prompting researchers to propose numerous subsequent improvements\cite{reza2004realization},\cite{wang2017image}. The Retinex algorithm\cite{land1983recent} simulates the color constancy of the human visual system by separating the illumination and reflection components of an image to enhance it. The Retinex model based on partial differential equations (\textit{\textit{PDE}}) \cite{osher2005iterative} constructs a PDE to differentiate between the illumination and reflection components of the image. In contrast, the variational method-based Retinex model\cite{palma2008perceptually},\cite{kimmel2003variational} effectively eliminates halo effects by optimizing the image's brightness and contrast. In the field of biomimetics, image enhancement methods based on visual cortical neural networks have also made significant progress. For instance, the Pulse-Coupled Neural Network (\textit{\textit{PCNN}})\cite{chen2011new} simulates the neural network structure of the mammalian visual cortex, enhancing images by capturing features and utilizing synchronized oscillation characteristics. Additionally, deep learning-based image enhancement methods have gained widespread application. Generative Adversarial Networks (\textit{\textit{GAN}})\cite{xu2022image} excel in image enhancement tasks through unsupervised or weakly supervised learning. Furthermore, multi-level feature fusion methods\cite{zamir2020learning} leverage convolutional neural networks to extract and combine multi-level image features, effectively improving the quality of low-light images. In\cite{wang2018phase}, an algorithm for image enhancement through specific thermal imaging steps is introduced to reduce diffusion effects. This method first applies a Fourier transform to the acquired thermal image, converting the image data from the spatiotemporal domain to the frequency domain. Subsequently, a deconvolution operation is performed to restore the pseudo-thermal flow, followed by phase-locking after reconversion to the temporal domain, effectively enhancing the contrast and clarity of the thermal image. In\cite{li2024blind}, a method for defocusing single-frame astronomical images is presented, which combines prior-weighted total variation and blind deconvolution techniques. This approach iteratively optimizes the skeleton image and the blur kernel, calculating the image Point Spread Function (\textit{\textit{PSF}}) and performing deconvolution to mitigate the decline in image quality caused by atmospheric turbulence distortions. 
Moreover, diffusion models\cite{croitoru2023diffusion},\cite{song2019generative} have shown remarkable performance in image enhancement. In the forward process, noise is gradually injected until the image is completely covered by noise. Subsequently, a trained neural network is employed to progressively restore the noisy image, removing noise frame by frame, ultimately generating or recovering a clearer image. 

In order to obtain recovered clear images from polarized images of foggy scenes, this paper proposes a defogging method based on the principle of physical diffusion. By collecting data on foggy days using a polarization camera, we simulate the diffusion of fog and apply the RPHF algorithm\cite{wang2018phase} to perform spatiotemporal Fourier transformation and deconvolution on the polarized images, thereby reducing the blurring effects caused by fog diffusion. Experimental results demonstrate that the proposed method outperforms several classic dehazing algorithms in terms of effectiveness and metrics, effectively restoring object contours and details, making it suitable for subsequent applications such as object detection and recognition. The remainder of this paper is organized as follows: Section~\ref{sec2} elaborates on the principles of the proposed method. Section~\ref{sec3} describes the experimental setup and data acquisition. Section~\ref{sec4} presents the experimental results and metric comparisons. Finally, Section~\ref{sec5} concludes the paper.

\section{Methodology}
\label{sec2}
\begin{figure}[!t]
\centering{\includegraphics[width=\columnwidth]{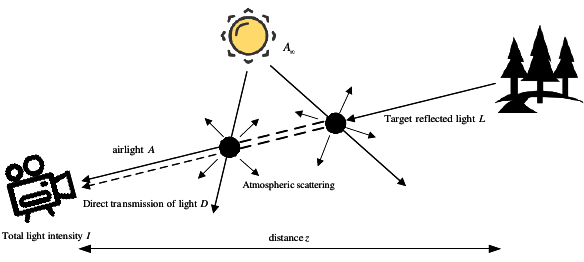}}
\caption{Atmospheric scattering model.}
\label{fig1}
\end{figure}

When there is a significant temperature difference between day and night and a large amount of moisture in the air, water vapor near the ground condenses into tiny liquid particles or ice crystals, suspended in the air to form fog. At this point, each liquid particle in the air acts like a micro-lens, continuously refracting and scattering light from objects as it passes through, resulting in a dimming effect on the light that reaches the human eye and the camera. Researchers have established atmospheric scattering models based on the principles of fog formation, simulating the scattering process of light by atmospheric particles as it reaches the camera. This modeling aids in the inversion of visible light during dehazing processes. Specifically, as shown in Fig.~\ref{fig1}, the light $I$ reaching the camera imaging system is primarily divided into two parts: direct transmission light $D$ and atmospheric light $A$. Direct transmission light $D$ is the scene radiance $L$ that has attenuated due to scattering, while atmospheric light $A$ consists of stray light caused by the scattering and refraction of numerous fog particles in the air\cite{narasimhan2000chromatic}. Since the target reflection light $L$ experiences significant reflection and refraction as it passes through fog particles, the transmission light $D$ at the camera imaging system exhibits an exponential decay with distance:
\begin{equation}I=D+A.\label{eq1}\end{equation}
\begin{equation}D = L e^{-\beta z}.\label{eq2}\end{equation}
where $\beta$ represents the attenuation coefficient, $z$ is the distance from the target to the lens,  $e^{-\beta z}$ defined as the transmittance.
We consider only the single scattering of haze particles here. The intensity of atmospheric light $A$ that reaches the imaging system increases exponentially with distance.
\begin{equation} A = A_{\infty} \left( 1 - e^{-\beta z} \right).\label{eq3}\end{equation}
where $A_{\infty}$ is the intensity of atmospheric light at an infinite distance.
Defining the transmittance as $t$, we can obtain the target reflected light by eliminating the direct transmitted light and the transmittance from the total light intensity:
\begin{equation} I = D + A = Lt + A_{\infty}(1 - t).\label{eq4}\end{equation}
\begin{equation}  L = \frac{I - A}{1-\frac{A}{A_{\infty}}} .\label{eq5}\end{equation}
Based on \eqref{eq5}, by estimating the atmospheric light intensity at infinity and the atmospheric light intensity at each pixel, we can perform inverse calculations to derive the target reflected light $L$.

During light transmission, the periodic regular changes in the direction of electric field vector oscillation lead to the phenomenon of polarization, which refers to the asymmetric spatial distribution of the light vector oscillation. The polarization imaging technology derived from this can detect polarized light generated by reflection, refraction, and scattering of light. The most used method to describe the polarization state of light is the Stokes vector method. This method uses four parameters, $S0$, $S1$, $S2$, and $S3$ to describe the polarized light:
\begin{equation} \mathbf{S} = 
\begin{pmatrix}
S_0 \\
S_1 \\
S_2 \\
S_3 \\
\end{pmatrix} = 
\begin{pmatrix}
I_0 + I_{90} \\
I_0 - I_{90} \\
I_{45} - I_{135} \\
I_r + I_l \\
\end{pmatrix}
\label{eq6}\end{equation}
where $I$ represents the polarization component at different angles, $I_r$ and $I_l$ correspond to right-hand circularly polarized light and left-hand circularly polarized light, respectively. $S0$ represents the total intensity of light, while the other Stokes parameters indicate the intensity difference between different polarization components. Based on the Stokes vector, the degree of linear polarization ($DOLP$) and angle of linear polarization ($AOLP$) of the target can be calculated:
\begin{equation} DOLP = \frac{\sqrt{S_1^2 + S_2^2}}{S_0}.\label{eq7}\end{equation}
\begin{equation} AOLP = \frac{1}{2} \arctan \frac{S_2}{S_1}.\label{eq8}\end{equation}

According to the Mie scattering model, the polarization characteristics of light scattered by haze particles under hazy weather conditions are minimally affected by the azimuth angle and exhibit partial polarization in all azimuth directions. Based on this scattering characteristic of haze particles, polarized optical imaging technology can be used to estimate atmospheric scattered light. Since the sky region does not contain any objects, the light intensity reaching the camera imaging system from the sky area only consists of atmospheric light and does not include reflected light from any objects. By extracting the sky region in the image and analyzing its polarization information, the atmospheric light intensity at infinity and the atmospheric light intensity $A$ at each pixel can be estimated, thereby achieving haze removal. 

The essence of fog diffusion is the movement of water vapor, which has a mathematical description similar to the diffusion of heat in a medium:
\begin{equation}  \frac{\partial C(\bm{X}, t)}{\partial t} = K \nabla^2 C(\bm{X}, t).\label{eq9}\end{equation}
where $C(\bm{X}, t)$ represents the fog concentration at position $\bm{X}(x, y, z)$, 
$K$ is the diffusion coefficient, $t$ represents the time of diffusion, 
and $\nabla^2$ is the Laplace operator. Based on this theory, let’s suppose that 
in a metaverse environment, fog is emitted from a certain point in the image 
and continues to diffuse over time, gradually producing an enhanced blurring effect:
\begin{equation} C(\bm{X}, 0) = \delta(\bm{X} - \bm{X}_0).\label{eq10}\end{equation}
where $\bm{X}_0$ represents the initial position, and $\delta$ represents the Dirac delta function. Therefore, a foggy image captured in the real world can be considered as the superposition of a clear background image and the diffused fog. In other words, the spatial diffusion of fog can be expressed as the convolution of the clear background image and the diffusion kernel:
\begin{equation} I_{ob}(u) = I_{bg}(u) * C(\bm{X}, t) + n.\label{eq11}\end{equation}
where $u$ is the depth of the image, $I_{ob}$ is the captured image, $I_{bg}$ is the clear background image, $*$ represents the convolution operation, and $n$ is the external noise. To simplify the problem, we assume that the light emitted by objects also follows a similar diffusion process, forming a continuous sequence. Under the condition of a semi-infinite space, the diffusion equation of fog typically has an analytical solution, which takes the form of a Gaussian distribution:
\begin{equation}  C(\bm{X}, t) = \frac{1}{\left(\sqrt{4 \pi K t}\right)^3} \exp \left( -\frac{(\bm{X} - \bm{X}_0)^2}{4 K t} \right).\label{eq12}\end{equation}
Applying the Fourier transform to \eqref{eq9} for both the time and spatial dimensions, we can easily obtain a frequency domain representation of the fog diffusion process:
\begin{equation}\widetilde{C}(\xi, w) = \frac{\widetilde{C}_0(\xi, w)}{\sqrt{\xi^2 + \frac{iw}{K}}}
\label{eq13}\end{equation}
where $\widetilde{C}(\xi, w)$ is the Fourier transform of the fog concentration, $\widetilde{C}_0(\xi, w)$ is the initial fog concentration, $\xi$ and $w$ represent the spatial and temporal frequencies, respectively. Equation (13) can describe how the Gaussian diffusion effect leads to the uneven distribution of fog concentration, affecting the degree of blurring in the image, especially the attenuation of high-frequency components.

Since it is difficult to obtain fog diffusion images at different times, we used Gaussian filters with varying radii to simulate the blurring effect caused by gradual diffusion over time, as shown in Fig. \ref{fig2}. According to Wang ’ s RPHF algorithm, we designed a deconvolution kernel $\sqrt{\xi^2 + i \omega / K}$ for the blurred images. This approach can mitigate the negative effects of Gaussian diffusion, thereby enhancing the details and contour information in the image:
\begin{figure*}[!t]
	\centering{\includegraphics[width=0.8\textwidth]{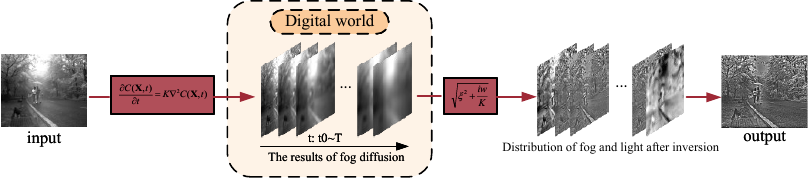}}
	\caption{Algorithm principle diagram.}
	\label{fig2}
\end{figure*}
\begin{equation}\tilde{C}_0(\xi, \omega) = \tilde{C}(\xi, \omega) \times \frac{1}{\tilde{\text{PSF}}(\xi, \omega)}.\label{eq14}\end{equation}
\begin{equation}\tilde{\text{PSF}}(\xi, \omega) = \frac{1}{\sqrt{\xi^2 + i \omega / K}}.\label{eq15}\end{equation}
where $\tilde{C}(\xi, \omega)$ represents the fog concentration at different times in the frequency domain, corresponding to the degree of blurring at different times in the image. We obtained the blur increment from the theoretically generated diffusion images, suggesting that the noise term $n$ in \eqref{eq11} can be ignored. In the image, high-frequency components represent edges and details, while diffusion weakens these components. Deconvolution enhances the weights of high-frequency components, restoring more details. By effectively reducing the diffusion effects of fog and light, the final result shows the state of fog points before diffusion and the pixel distribution without light sources.

\section{Experimental setup and data acquisition}
\label{sec3}
In this section, we present the experimental details in Part~\ref{part A}, while Part~\ref{part B} provides a detailed description of the data acquisition and its sources.

\subsection{Experimental Details}
\label{part A}
We first replicate the input Stokes vector image into 100 layers to simulate images at different time points. Then, Gaussian filtering is applied to each frame, with the filter radius increasing over time. The data is then downsampled and cropped along the time axis. By subtracting each time point from the original image and filtering to remove abrupt changes, we obtain the blur increment. The increment is then 3D-padded using edge replication and spatially downsampled. To slow and smooth the fog diffusion, we apply temporal filtering for each pixel and extend the time axis to prevent frequency leakage before performing spatio-temporal Fourier transforms.

In the frequency domain processing, we first generate a 2D-frequency matrix  $\xi^2$, based on the current image dimensions, and a temporal frequency matrix $\omega$ , based on the extended time axis. We set the diffusion coefficient to 1 and compute the deconvolution kernel for the image. Then, we multiply the Fourier transform result by the deconvolution kernel and convert it back to the spatial domain. Finally, to smooth the data, we average adjacent layers along the time dimension. The output is then resized through upsampling and downsampling, ensuring that the output dimensions match the input image. Each input image generates 51 output images, and after cropping, the size of the output images matches the original input images. The series of output images is then averaged and inverted to produce the final output image. 

\subsection{Dataset}
\label{part B}
\subsubsection{Real Dataset }
We used a polarization camera to capture polarization information in outdoor foggy scenes. During the experiment, the polarization camera was connected to a computer, ensuring that the camera's position remained relatively fixed during each image capture, with the optical axis kept parallel to the horizontal plane. To ensure flexibility and stability in the data collection process, we used a mobile tripod and a gimbal to adjust and secure the camera’s position, while a portable power source supplied power to both the camera and the computer.

The polarization camera used in the experiment is a Blackfly polarization camera based on the Sony IMX250MZR sensor. This camera is capable of real-time measurement of the polarization patterns in a scene with a single exposure, capturing the polarization information of various targets in the scene. Since the image data output by the Blackfly polarization camera is in raw format, which represents the raw data of light signals converted to digital signals by the sensor, it cannot be used directly as an image. Therefore, it is necessary to process the data to obtain Stokes vector images, degree of polarization images, and polarization angle images.

We used the SDK software provided with the Blackfly polarization camera to process the raw polarization images. Each computational unit in the image consists of four adjacent pixels, corresponding to polarization-sensitive directions of 0°, 45°, 90°, and 135°. By extracting the pixels with the same polarization-sensitive direction from each unit, we obtained separate polarization component images for 0°, 45°, 90°, and 135°. The brightness of each polarization component image is half of the original, and the resolution is 1/4 of the original (1224$\times$1024).

\begin{figure}[!t]
	\centering{\includegraphics[width=\columnwidth]{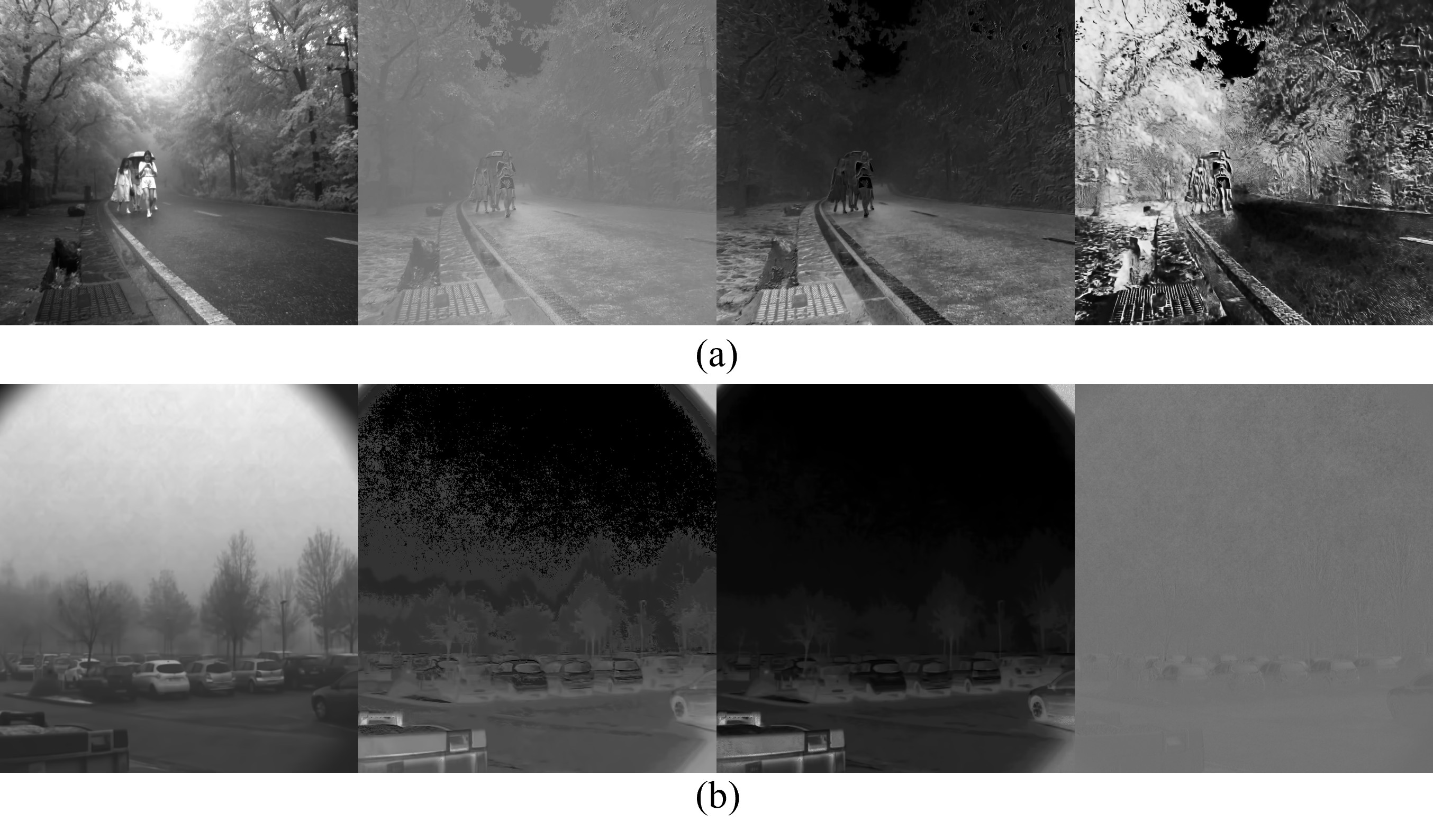}}
	\caption{The datasets we used are as follows: (a) is the real dataset we collected, with $S0$, $S1$, $DOLP$ and $AOLP$, images of the same set arranged from left to right. (b) is the open-source dataset.
	}
	\label{fig3}
\end{figure}

Using the Stokes vector calculation formulas, we obtained the first three components of the Stokes vector, $S0$, $S1$ and $S2$, with image resolutions consistent with the polarization component images at each angle. Additionally, we calculated the degree of linear polarization ($DOLP$) and the angle of linear polarization ($AOLP$) images. In total, we collected 21 sets of polarization images in outdoor foggy scenes, and the corresponding $S0$, $S1$, $DOLP$ and $AOLP$ images were calculated, as shown in Fig.~\ref{fig3}(a).

\subsubsection{Public Dataset }
The PolarLITIS Dataset\cite{blin2021polarlitis} is the first multimodal dataset containing paired RGB and polarization images. It includes scenes from highways, urban areas, rural settings, parking lots, and campuses, covering weather conditions such as sunny, cloudy, and hazy. Multiple planned data formats are provided. We selected 509 Stokes images from the test set, processed them to obtain $S0$, $S1$, $DOLP$, and $AOLP$ images, each with a resolution of 500×500 pixels, as shown in Fig.~\ref{fig3}(b).

\section{Results}
\label{sec4}
This section provides qualitative and quantitative analyses of dehazing results across different methods on both real and public datasets, along with comparisons of selected experimental images and metrics. Our results were compared on $S0$ and $DOLP$ with Histogram Equalization, Retinex\cite{parthasarathy2012automated}, Dark Channel\cite{he2010single}, Polarization\cite{liang2014method}, and DehazeNet\cite{cai2016dehazenet} methods. To provide a more intuitive comparison of the dehazing effects, we performed unified histogram matching on all outputs using the original images as references.

\begin{figure*}[!t]
	\centering
	\includegraphics[width=\textwidth]{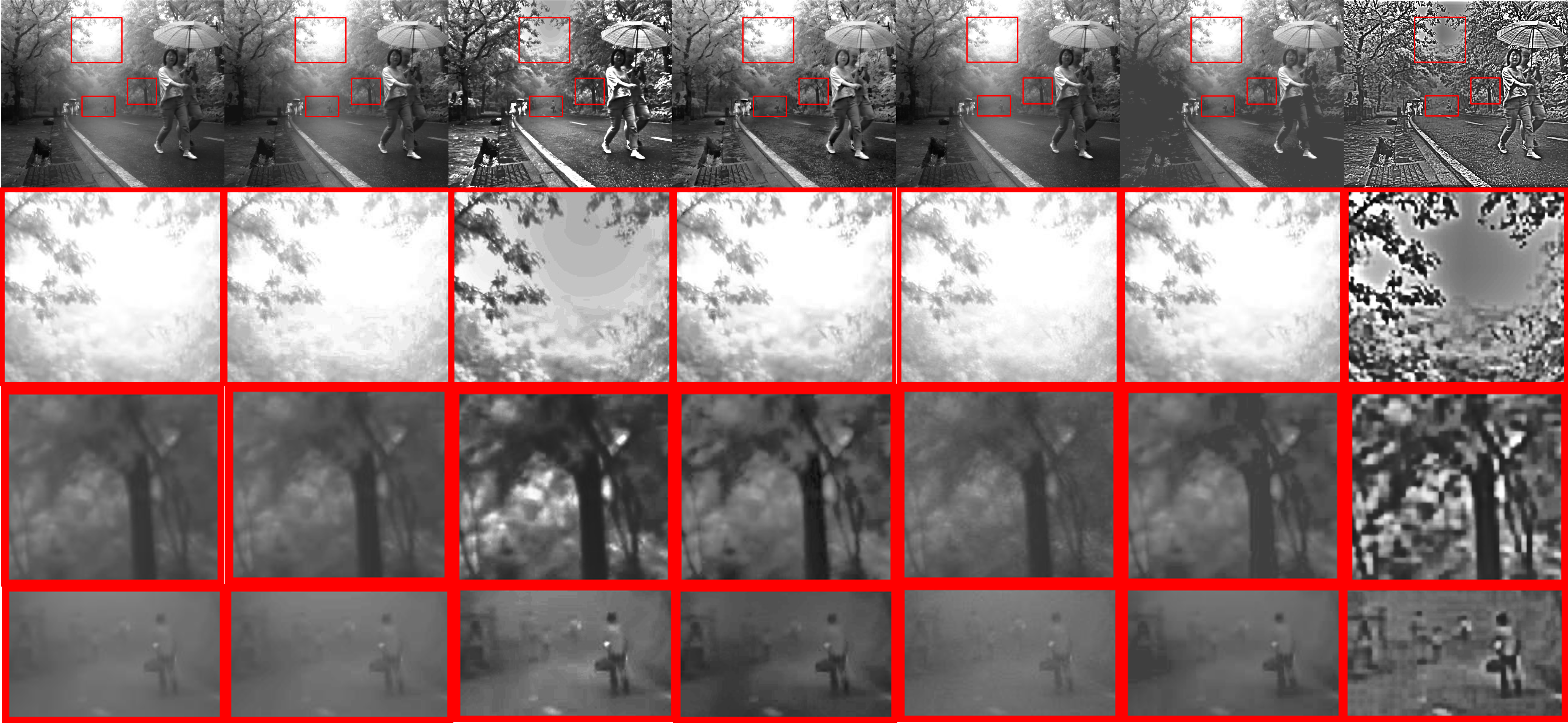} 
\vspace{0.5em} 
\noindent
\makebox[\textwidth][c]{%
	\parbox{0.13\textwidth}{\centering \fontsize{7pt}{10pt}\selectfont (a) Original scene} \hfill
	\parbox{0.13\textwidth}{\centering \fontsize{7pt}{10pt}\selectfont (b) Histeq} \hfill
	\parbox{0.13\textwidth}{\centering \fontsize{7pt}{10pt}\selectfont (c) Retinex} \hfill
	\parbox{0.13\textwidth}{\centering \fontsize{7pt}{10pt}\selectfont (d) Dark-Channel} \hfill
	\parbox{0.13\textwidth}{\centering \fontsize{7pt}{10pt}\selectfont (e) Polarization} \hfill
	\parbox{0.13\textwidth}{\centering \fontsize{7pt}{10pt}\selectfont (f) DehazeNet} \hfill
	\parbox{0.13\textwidth}{\centering \fontsize{7pt}{10pt}\selectfont (g) Proposed}
}
	\caption{Dehazing results for $S0$ images in the real dataset.}
	\label{fig4}
\end{figure*}

\begin{figure*}[!t]
	\centering
	\includegraphics[width=\textwidth]{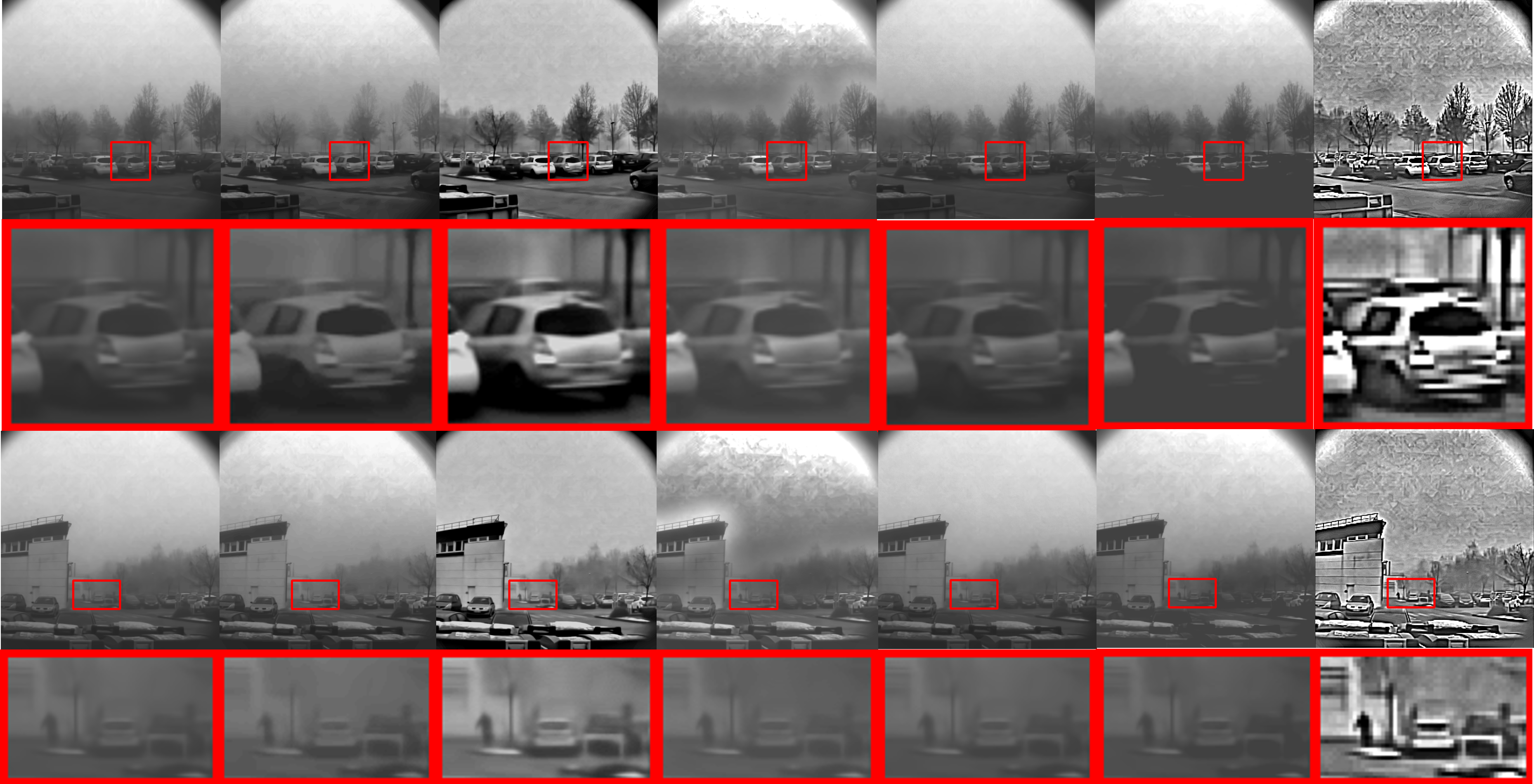} 
	\vspace{0.5em} 
	\noindent
	\makebox[\textwidth][c]{%
		\parbox{0.13\textwidth}{\centering \fontsize{7pt}{10pt}\selectfont (a) Original scene} \hfill
		\parbox{0.13\textwidth}{\centering \fontsize{7pt}{10pt}\selectfont (b) Histeq} \hfill
		\parbox{0.13\textwidth}{\centering \fontsize{7pt}{10pt}\selectfont (c) Retinex} \hfill
		\parbox{0.13\textwidth}{\centering \fontsize{7pt}{10pt}\selectfont (d) Dark-Channel} \hfill
		\parbox{0.13\textwidth}{\centering \fontsize{7pt}{10pt}\selectfont (e) Polarization} \hfill
		\parbox{0.13\textwidth}{\centering \fontsize{7pt}{10pt}\selectfont (f) DehazeNet} \hfill
		\parbox{0.13\textwidth}{\centering \fontsize{7pt}{10pt}\selectfont (g) Proposed}
	}
	\caption{Dehazing results for $S0$ images in the public dataset.}
	\label{fig5}
\end{figure*}

\begin{figure*}[!t]
	\centering
	\includegraphics[width=\textwidth]{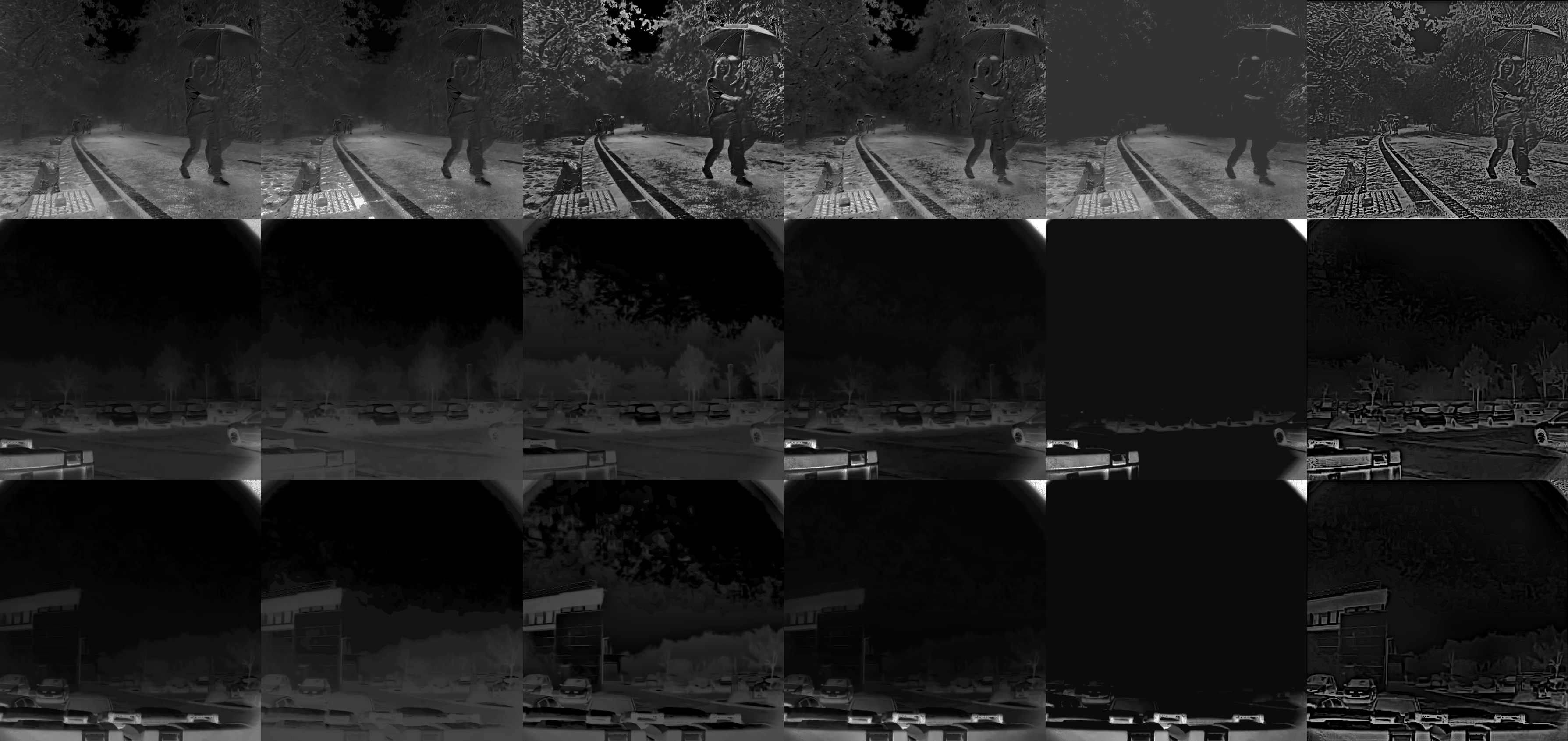} 
	\vspace{0.5em} 
	\noindent
	\makebox[\textwidth][c]{%
		\parbox{0.13\textwidth}{\centering \fontsize{7pt}{10pt}\selectfont (a) Original scene} \hfill
		\parbox{0.13\textwidth}{\centering \fontsize{7pt}{10pt}\selectfont (b) Histeq} \hfill
		\parbox{0.13\textwidth}{\centering \fontsize{7pt}{10pt}\selectfont (c) Retinex} \hfill
		\parbox{0.13\textwidth}{\centering \fontsize{7pt}{10pt}\selectfont (d) Dark-Channel} \hfill
		\parbox{0.13\textwidth}{\centering \fontsize{7pt}{10pt}\selectfont (e) DehazeNet} \hfill
		\parbox{0.13\textwidth}{\centering \fontsize{7pt}{10pt}\selectfont (f) Proposed}
	}
	\caption{Dehazing results for $DOLP$ images.}
	\label{fig6}
\end{figure*}

Fig.~\ref{fig4} and Fig.~\ref{fig5} show the $S0$ image results in different scenes. Histogram equalization only slightly enhanced the image contrast. The Retinex performed well in our experiments, significantly enhancing contrast, but it produced an odd halo effect in the sky regions of the real dataset. The Dark Channel generated images with more uniform grayscale values, such as clearer facial contours under an umbrella. However, it also created a white halo on the roof of the building in Fig.~\ref{fig5}. The Polarization and DehazeNet performed poorly in this experiment, with limited dehazing effects. In particular, DehazeNet caused excessive darkening and blurring in shadowed areas. In contrast, since our algorithm restores the state before fog diffusion as well as the distribution after light inversion, the results are obtained without illumination and with “lights off”, which shows that we are able to harmonize the pixel grayscale of the light and dark areas well. So our algorithm can clearly revealed the contours of objects in the images and effectively removed the detail blurring caused by haze in the original images. For example, in Fig.~\ref{fig4},  our algorithm clearly shows the outline of the face under the umbrella in the near distance, the shape of the tree canopy and the person in the far distance, as well as being able to clearly distinguish the lamp post from the tree trunk. At the same time, our algorithm makes the transparent umbrella held in the right hand of the person in the distance clearer. In Fig.~\ref{fig5}, the tires on the underside of the vehicle and the dent formed by the impact at the rear become more visible, and the streaks on the far wall become more pronounced, although there are unsmooth edges in these localized areas. The dehazing effects on the $DOLP$ images are shown in Fig.~\ref{fig6}. It can be seen that our algorithm effectively enhances the detailed parts but does not produce excessive noise like Retinex. Overall, our algorithm provides better enhancement for both nearby and distant areas, making it well-suited for subsequent tasks such as object detection and segmentation.

Additionally, we conducted a quantitative analysis of the dehazing results. Since no clear ground truth images are available, we applied no-reference image quality assessment metrics $e$, $\overline{r}$, $\sigma$ \cite{hautiere2008blind}, the standard deviation and the average gradient for the dehazing of polarization images. Where the definitions of $e$, $\overline{r}$, and $\sigma$ are as follows:

\subsubsection{$e$: Rate of new visible edges}
\begin{equation}e = \frac{n_r - n_o}{n_o}.\label{eq16}\end{equation}
where $n_r$ and $n_o$ represent the number of visible edges in the restored image and the original foggy image, respectively. A higher $e$ value indicates better edge visibility and sharper details.

\subsubsection{$\overline{r}$: Degree of gradient enhancement}
\begin{equation}
	\overline{r} = \exp \left( \frac{1}{n_r} \sum_{P_i \in \varTheta_r} \log r_i \right). 
	\label{eq17}
\end{equation}
where $\varTheta_r$ is the set of visible edges in the restored image, $P_i$ represents an individual edge in the set, and $r$ is the gradient ratio between the restored and original images. A higher $\overline{r}$ indicates better preservation of texture information.

\subsubsection{$\sigma$: Rate of new visible edges}
\begin{equation}
\sigma = \frac{n_s}{\mathrm{dim}_x \times \mathrm{dim}_y} \label{eq18}
\end{equation}
where $n_s$ represents the number of pixels that became saturated after restoration but were not saturated before, and the denominator is the image size. A lower $\sigma$ indicates less information loss and better dehazing results.

We calculated five metrics for different dehazing algorithms on $S0$ and $DOLP$, as well as the metrics after unification, as shown in Table~\ref{tab 1} and Table~\ref{tab 2}. The data marked with “$*$” represents the results after histogram matching. The best dehazing results are highlighted in blue, while the best results after unification are highlighted in red. For the most part, our method has better values before and after histogram matching, with the first two and the average gradient metrics for polarization information significantly higher than other methods, and the third metric also performs well. This indicates that our method performs superiorly in enhancing edges and gradients. Our algorithm performs poorly in terms of standard deviation, which is somewhat justified: because our results do not include the distribution of light, reducing the dynamic range of gray values, but this effect is eliminated after histogram matching, and these metrics are significantly improved. Overall, the proposed method performs best in most cases.

\begin{table*}[ht]
	\centering
\caption{Comparison of Result Metrics for Real Dataset}
\label{tab 1}
\renewcommand{\arraystretch}{1.2} 
\setlength{\tabcolsep}{4pt} 
\begin{tabular}{lccccc|ccccc} 
	\toprule
	\multicolumn{1}{c}{} & \multicolumn{5}{c}{\textbf{S0}} & \multicolumn{5}{c}{\textbf{DOLP}}\\
	\cmidrule(r){2-6} \cmidrule(r){7-11}
		\textbf{} & $e \uparrow$ & $\overline{r} \uparrow$ & $\sigma \downarrow$ & $SD \uparrow$ & $AG \uparrow$ & $e \uparrow$ & $\overline{r} \uparrow$ & $\sigma \downarrow$ & $SD \uparrow$ & $AG \uparrow$ \\
		\midrule
		\textbf{Histeq} & 0.536398 & 1.785461 & 0.014087 & \textcolor{blue}{74.909897} & 41.149748 & 0.958568 & 3.473655 & 0.011153 & \textcolor{blue}{74.707190} & 54.950645 \\
		\textbf{Retinex} & 0.575552 & 0.691593 & \textcolor{blue}{0.000001} & 14.091436 & 17.097656 & 0.245433 & 1.019103 & \textcolor{blue}{0.000001} & 19.405553 & 18.606977 \\
		\textbf{Dark Channel} & 0.094412 & 1.438942 & 0.002856 & 55.731467 & 35.599269 & 0.115539 & 1.500083 & 0.006108 & 34.332712 & 33.155088 \\
		\textbf{Polarization} & 0.278274 & 1.214958 & 0.002727 & 57.311402 & 33.808948 & -- & -- & -- & -- & -- \\
		\textbf{DehazeNet} & 0.075787 & 1.411419 & 0.326516 & 63.580747 & 32.686241 & 0.101919 & 1.169579 & 0.547953 & 28.563358 & 20.227125 \\
		\textbf{Proposed} & \textcolor{blue}{1.261295} & \textcolor{blue}{3.641936} & 0.000007 & 30.886114 & \textcolor{blue}{70.959909} &\textcolor{blue}{1.392921} & \textcolor{blue}{4.205956} & 0.000002 & 26.742624 & \textcolor{blue}{65.111840} \\
		\midrule
		\textbf{Histeq*} & 0.093663 & 1.095003 & 0.000798 & 57.424578 & 30.581238 & 0.087384 & 1.286197 & 0.003759 & \textcolor{red}{40.807419} & 29.301364 \\
		\textbf{Retinex*} & 1.016016 & 3.372110 & 0.008316 & \textcolor{red}{58.206270} & 78.925608 & 1.137773 & 3.891999 & 0.003789 & 39.101555 & 67.747182 \\
		\textbf{Dark Channel*} & 0.542571 & 1.848603 & 0.000271 & 56.774233 & 47.983325 & 0.594967 & 1.872890 & 0.002790 & 38.025393 & 37.726988\\
		\textbf{Polarization*} & 0.283390 & 1.214129 & 0.002055 & 57.186545 & 33.715011 & -- & -- & -- & -- & -- \\
		\textbf{DehazeNet*} & 0.196265 & 1.161530 & 0.000209 & 51.298981 & 28.655317 & 0.238613 & 1.193448 & \textcolor{red}{0.000001} & 31.199593 & 20.716168 \\
		\textbf{Proposed*} & \textcolor{red}{1.776061} & \textcolor{red}{7.558234} & \textcolor{red}{0.000004} & 57.310984 & \textcolor{red}{130.793137} & \textcolor{red}{1.760050} & \textcolor{red}{7.962545} & 0.024138 & 38.149934 & \textcolor{red}{93.401123} \\
		\bottomrule
	\end{tabular}
\end{table*}

\begin{table*}[ht]
	\centering
	\caption{Comparison of Result Metrics for Public Dataset}
	\label{tab 2}
	\renewcommand{\arraystretch}{1.2} 
	\setlength{\tabcolsep}{4pt} 
	\begin{tabular}{lccccc|ccccc} 
	\toprule
		\multicolumn{1}{c}{} & \multicolumn{5}{c}{\textbf{S0}} & \multicolumn{5}{c}{\textbf{DOLP}} \\
		\cmidrule(r){2-6} \cmidrule(r){7-11}
		\textbf{} & $e \uparrow$ & $\overline{r} \uparrow$ & $\sigma \downarrow$ & $SD \uparrow$ & $AG \uparrow$ & $e \uparrow$ & $\overline{r} \uparrow$ & $\sigma \downarrow$ & $SD \uparrow$ & $AG \uparrow$ \\
		\midrule
		\textbf{Histeq} & 1.129682 & 2.389591 & 0.030348 & \textcolor{blue}{74.822240} & 22.266143 & 0.890653 & 3.168015 & 0.017225 & \textcolor{blue}{68.325060} & 21.302955 \\
		\textbf{Retinex} & \textcolor{blue}{1.523904} & 1.496992 & \textcolor{blue}{0.000001} & 25.406687 & 16.507464 & \textcolor{blue}{1.133409} & 2.427125 & \textcolor{blue}{0.000001} & 26.619163 & 19.220979 \\
		\textbf{Dark Channel} & 0.400953 & 2.151798 & 0.004613 & 44.826645 & 18.973456 & 0.003636 & 0.939801 & 0.179782 & 24.485523 & 13.784155 \\
		\textbf{Polarization} & 0.374919 & 1.978670 & \textcolor{blue}{0.000001} & 58.942494 & 16.068535 & -- & -- & -- & -- & -- \\
		\textbf{DehazeNet} & 0.093171 & 1.327234 & 0.129968  & 63.765027 & 16.499216 & 0.005125 & 1.037528 & 0.725762 & 21.791114 & 10.388810 \\
		\textbf{Proposed} & 1.191910 & \textcolor{blue}{8.626697} & \textcolor{blue}{0.000001} & 20.737691 & \textcolor{blue}{44.134689} & 0.857636 & \textcolor{blue}{8.675575} & \textcolor{blue}{0.000001} & 23.568484 & \textcolor{blue}{41.159597} \\
		\midrule
		\textbf{Histeq*} & 1.196586 & 1.545085 & 0.004873 & 54.705322 & 16.478460 & 0.816598 & 1.101802 & 0.228058 & 26.348177 & 8.623575 \\
		\textbf{Retinex*} & 1.543148 & 3.041670 & 0.024821 & 53.434641 & 34.898098 & \textcolor{red}{1.141202} & 2.111275 & 0.116897 & 25.434418 & 18.565283 \\
		\textbf{Dark Channel*} & 0.366791 & 2.004151 & 0.051418 & 48.455792 & 20.452834 & 0.006267 & 1.171254 & \textcolor{red}{0.001276} & \textcolor{red}{30.382642} & 17.044026 \\
		\textbf{Polarization*} & 0.064327 & 1.043226 & \textcolor{red}{0.000001} & \textcolor{red}{54.847559} & 13.959261 & -- & -- & -- & -- & -- \\
		\textbf{DehazeNet*} & 0.131112 & 1.142698 & 0.008960 & 54.558325 & 13.992031 & 0.014428 & 1.456572 & 0.001942 & 30.117631 & 14.142568 \\
		\textbf{Proposed*} & \textcolor{red}{2.516799} & \textcolor{red}{36.521237} & 0.051273 & 49.143453 & \textcolor{red}{103.822755} & 0.899406 & \textcolor{red}{9.324635} & 0.087051 & 27.687805 & \textcolor{red}{43.662665} \\
		\bottomrule
	\end{tabular}
\end{table*}

\section{Conclusion}
\label{sec5}
This paper presents a polarization image enhancement algorithm designed to restore details from deep blur. The algorithm simulates real-world fog diffusion by generating multi-scale images and designs a deconvolution kernel based on the corresponding point spread function, processing and cropping images in both the time and frequency domains to produce a series of enhanced images. The experiments utilized haze images captured with a polarization camera and public datasets. Experimental results demonstrate that the proposed method outperforms other classic dehazing algorithms in both qualitative and quantitative evaluations, effectively revealing target contours and scene details in hazy environments. Since we have not yet rendered the enhanced results, the most intuitive dehazing effects have not been fully realized. Future work will focus on optimizing the algorithm and developing suitable rendering methods to minimize image distortion, thus expanding the applicability of this approach to broader image processing tasks.

\bibliographystyle{ieeetr}
\bibliography{reference.bib}

\begin{thebibliography}{10}

\bibitem{wolff1997polarization}
L.~B. Wolff, ``Polarization vision: a new sensory approach to image
  understanding,'' {\em Image and Vision computing}, vol.~15, no.~2,
  pp.~81--93, 1997.

\bibitem{atkinson2021polarized}
G.~A. Atkinson, ``Polarized light in computer vision,'' in {\em Computer
  Vision: A Reference Guide}, pp.~1005--1010, Springer, 2021.

\bibitem{schechner2003polarization}
Y.~Y. Schechner, S.~G. Narasimhan, and S.~K. Nayar, ``Polarization-based vision
  through haze,'' {\em Applied optics}, vol.~42, no.~3, pp.~511--525, 2003.

\bibitem{narasimhan2000chromatic}
S.~G. Narasimhan and S.~K. Nayar, ``Chromatic framework for vision in bad
  weather,'' in {\em Proceedings IEEE Conference on Computer Vision and Pattern
  Recognition. CVPR 2000 (Cat. No. PR00662)}, vol.~1, pp.~598--605, IEEE, 2000.

\bibitem{schechner2001instant}
Y.~Y. Schechner, S.~G. Narasimhan, and S.~K. Nayar, ``Instant dehazing of
  images using polarization,'' in {\em Proceedings of the 2001 IEEE Computer
  Society Conference on Computer Vision and Pattern Recognition. CVPR 2001},
  vol.~1, pp.~I--I, IEEE, 2001.

\bibitem{zhang2021review}
W.~Zhang, J.~Liang, G.~Wang, H.~Zhang, and S.~Fu, ``Review of passive
  polarimetric dehazing methods,'' {\em Optical Engineering}, vol.~60, no.~3,
  pp.~030901--030901, 2021.

\bibitem{liang2014method}
J.~Liang, L.~Ren, E.~Qu, B.~Hu, and Y.~Wang, ``Method for enhancing visibility
  of hazy images based on polarimetric imaging,'' {\em Photonics Research},
  vol.~2, no.~1, pp.~38--44, 2014.

\bibitem{namer2005advanced}
E.~Namer and Y.~Y. Schechner, ``Advanced visibility improvement based on
  polarization filtered images,'' in {\em Polarization Science and Remote
  Sensing II}, vol.~5888, pp.~36--45, SPIE, 2005.

\bibitem{xu2023research}
L.~Xu and Y.~Li, ``Research on underwater imaging technology based on
  polarization difference,'' {\em Journal of Optics}, pp.~1--8, 2023.

\bibitem{hu2018polarimetric}
H.~Hu, L.~Zhao, X.~Li, H.~Wang, J.~Yang, K.~Li, and T.~Liu, ``Polarimetric
  image recovery in turbid media employing circularly polarized light,'' {\em
  Optics Express}, vol.~26, no.~19, pp.~25047--25059, 2018.

\bibitem{nithyananda2016review}
C.~Nithyananda, A.~Ramachandra, {\em et~al.}, ``Review on histogram
  equalization based image enhancement techniques,'' in {\em 2016 International
  Conference on Electrical, Electronics, and Optimization Techniques (ICEEOT)},
  pp.~2512--2517, IEEE, 2016.

\bibitem{reza2004realization}
A.~M. Reza, ``Realization of the contrast limited adaptive histogram
  equalization (clahe) for real-time image enhancement,'' {\em Journal of VLSI
  signal processing systems for signal, image and video technology}, vol.~38,
  pp.~35--44, 2004.

\bibitem{wang2017image}
Y.~Wang and Z.~Pan, ``Image contrast enhancement using adjacent-blocks-based
  modification for local histogram equalization,'' {\em Infrared Physics \&
  Technology}, vol.~86, pp.~59--65, 2017.

\bibitem{land1983recent}
E.~H. Land, ``Recent advances in retinex theory and some implications for
  cortical computations: color vision and the natural image.,'' {\em
  Proceedings of the National Academy of Sciences}, vol.~80, no.~16,
  pp.~5163--5169, 1983.

\bibitem{osher2005iterative}
S.~Osher, M.~Burger, D.~Goldfarb, J.~Xu, and W.~Yin, ``An iterative
  regularization method for total variation-based image restoration,'' {\em
  Multiscale Modeling \& Simulation}, vol.~4, no.~2, pp.~460--489, 2005.

\bibitem{palma2008perceptually}
R.~Palma-Amestoy, E.~Provenzi, M.~Bertalm{\'\i}o, and V.~Caselles, ``A
  perceptually inspired variational framework for color enhancement,'' {\em
  IEEE transactions on pattern analysis and machine intelligence}, vol.~31,
  no.~3, pp.~458--474, 2008.

\bibitem{kimmel2003variational}
R.~Kimmel, M.~Elad, D.~Shaked, R.~Keshet, and I.~Sobel, ``A variational
  framework for retinex,'' {\em International Journal of computer vision},
  vol.~52, pp.~7--23, 2003.

\bibitem{chen2011new}
Y.~Chen, S.-K. Park, Y.~Ma, and R.~Ala, ``A new automatic parameter setting
  method of a simplified pcnn for image segmentation,'' {\em IEEE transactions
  on neural networks}, vol.~22, no.~6, pp.~880--892, 2011.

\bibitem{xu2022image}
B.~Xu, D.~Zhou, and W.~Li, ``Image enhancement algorithm based on gan neural
  network,'' {\em IEEE Access}, vol.~10, pp.~36766--36777, 2022.

\bibitem{zamir2020learning}
S.~W. Zamir, A.~Arora, S.~Khan, M.~Hayat, F.~S. Khan, M.-H. Yang, and L.~Shao,
  ``Learning enriched features for real image restoration and enhancement,'' in
  {\em Computer Vision--ECCV 2020: 16th European Conference, Glasgow, UK,
  August 23--28, 2020, Proceedings, Part XXV 16}, pp.~492--511, Springer, 2020.

\bibitem{wang2018phase}
H.~Wang, N.~Wang, Z.~He, and Y.~He, ``Phase-locked restored pseudo heat flux
  thermography for detecting delamination inside carbon fiber reinforced
  composites,'' {\em ieee transactions on industrial informatics}, vol.~15,
  no.~5, pp.~2938--2946, 2018.

\bibitem{li2024blind}
S.~Li, Y.~Gao, J.~Wu, and L.~Cao, ``Blind deblurring of astronomical images
  using a scgtv-based single-frame method,'' {\em Optics Express}, vol.~32,
  no.~20, pp.~35579--35593, 2024.

\bibitem{croitoru2023diffusion}
F.-A. Croitoru, V.~Hondru, R.~T. Ionescu, and M.~Shah, ``Diffusion models in
  vision: A survey,'' {\em IEEE Transactions on Pattern Analysis and Machine
  Intelligence}, vol.~45, no.~9, pp.~10850--10869, 2023.

\bibitem{song2019generative}
Y.~Song and S.~Ermon, ``Generative modeling by estimating gradients of the data
  distribution,'' {\em Advances in neural information processing systems},
  vol.~32, 2019.

\bibitem{blin2021polarlitis}
R.~Blin, S.~Ainouz, S.~Canu, and F.~Meriaudeau, ``The polarlitis dataset: Road
  scenes under fog,'' {\em IEEE transactions on intelligent transportation
  systems}, vol.~23, no.~8, pp.~10753--10762, 2021.

\bibitem{parthasarathy2012automated}
S.~Parthasarathy and P.~Sankaran, ``An automated multi scale retinex with color
  restoration for image enhancement,'' in {\em 2012 National Conference on
  Communications (NCC)}, pp.~1--5, IEEE, 2012.

\bibitem{he2010single}
K.~He, J.~Sun, and X.~Tang, ``Single image haze removal using dark channel
  prior,'' {\em IEEE transactions on pattern analysis and machine
  intelligence}, vol.~33, no.~12, pp.~2341--2353, 2010.

\bibitem{cai2016dehazenet}
B.~Cai, X.~Xu, K.~Jia, C.~Qing, and D.~Tao, ``Dehazenet: An end-to-end system
  for single image haze removal,'' {\em IEEE transactions on image processing},
  vol.~25, no.~11, pp.~5187--5198, 2016.

\bibitem{hautiere2008blind}
N.~Hautiere, J.-P. Tarel, D.~Aubert, and E.~Dumont, ``Blind contrast
  enhancement assessment by gradient ratioing at visible edges,'' {\em Image
  Analysis and Stereology}, vol.~27, no.~2, pp.~87--95, 2008.

\end{thebibliography}
\end{document}